\documentclass[letterpaper, 10 pt, conference]{ieeeconf}  % Comment this line out if you need a4paper

\IEEEoverridecommandlockouts                              % This command is only needed if 
\usepackage{graphics} % for pdf, bitmapped graphics files
\usepackage{epsfig} % for postscript graphics files
\usepackage{mathptmx} % assumes new font selection scheme installed
\usepackage{times} % assumes new font selection scheme installed
\usepackage{amsmath} % assumes amsmath package installed
\usepackage{amssymb}  % assumes amsmath package installed
\usepackage{tabularx}
\usepackage[ruled,vlined]{algorithm2e}
\usepackage{booktabs,multirow,makecell,subcaption}
\usepackage{tabularx}
\usepackage{placeins}

\title{\LARGE \bf
FSAG: Enhancing Human-to-Dexterous-Hand Finger-Specific Affordance Grounding via Diffusion Models
}

 \author{
Yifan Han$^{1,2}$,
Yichuan Peng$^{5}$,
Pengfei Yi$^{1,2}$,
Junyan Li$^{1,2}$,
Hanqing Wang$^{3}$,\\
Gaojing Zhang$^{4}$,
Qi~Peng~Liu$^{5}$,
Wenzhao~Lian$^{5\dagger}$%
\thanks{$^{1}$Institute of Automation, Chinese Academy of Sciences.;\;
$^{2}$School of Artificial Intelligence, University of Chinese Academy of Sciences.;\;
$^{3}$The Hong Kong University of Science and Technology (Guangzhou).;\;
$^{4}$University of Sussex.;\;
$^{5}$School of Artificial Intelligence, Shanghai Jiao Tong University. \texttt{lianwenzhao@sjtu.edu.cn}.;\;
$^{\dagger}$Corresponding Author.}%
}

\begin{document}

\setcounter{figure}{1}
\makeatletter
\let\@oldmaketitle=\@maketitle % store @maketitle
\renewcommand{\@maketitle}{\@oldmaketitle
\begin{center}
\includegraphics[width=\textwidth]{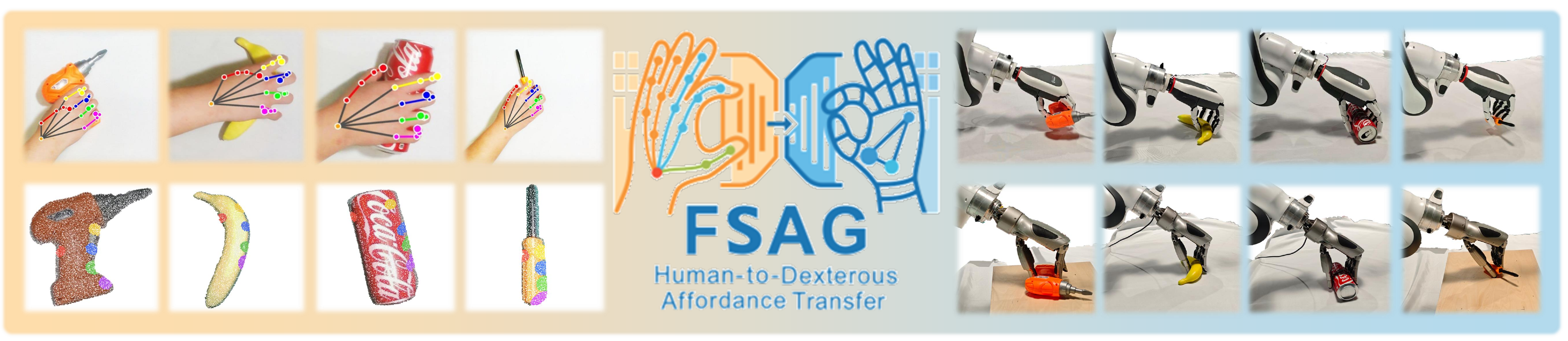}
\label{fig:cover}
\end{center}
\vspace{-15pt}
\footnotesize{Fig.~\thefigure.~\label{fig:cover} \textbf{Overview.} FSAG learns finger-specific affordance fields from human demonstrations by leveraging semantic priors of hand-object interactions from a frozen Stable Diffusion model~\cite{rombach2022high}. by conditioning grasp synthesis on affordance field learned from human demonstrations, our method generates human-intuitive and physically stable grasps, without requiring any robot action data.}
\vspace{-15pt}
\medskip}
\makeatother
\maketitle
\thispagestyle{empty}
\pagestyle{empty}

\maketitle
\thispagestyle{empty}
\pagestyle{empty}

% \begin{document}

% \makeatletter
% \let\@oldmaketitle=\@maketitle
% \renewcommand{\@maketitle}{\@oldmaketitle
% \begin{center}
% \includegraphics[width=\textwidth]{figs/logo.pdf}
% \end{center}

% \refstepcounter{figure}\label{fig:cover}
% \footnotesize{\figurename~\thefigure.~\textbf{Overview.}
% FSAG learns finger-specific affordance fields from human demonstrations by leveraging hand–object interaction semantic priors from a frozen Stable Diffusion~\cite{rombach2022high}. The predicted per-finger contact maps are fused with depth/segmentation to project contacts onto 3D geometry and drive an approach–closure–hold motion planner. Requiring no robot action data, FSAG yields functional, stable grasps on everyday objects and tools and transfers across dexterous-hand embodiments.}
% \vspace{-15pt}
% \medskip}
% \makeatother

% \maketitle
% \thispagestyle{empty}
% \pagestyle{empty}

%%%%%%%%%%%%%%%%%%%%%%%%%%%%%%%%%%%%%%%%%%%%%%%%%%%%%%%%%%%%%%%%%%%%%%%%%%%%%%%%
%灵巧手抓取一直是灵巧手操作的核心问题，由于更高自由度的问题，夹爪抓取的算法很难直接迁移到灵巧手上。现在的大部分方法依赖于大规模的灵巧手仿真或者现实中的抓取数据，并且难以迁移到不同的灵巧手上，随着新的灵巧手产品的不断诞生，过去的数据集已经无法满足现在的需求，而针对每一种灵巧手制作数据需要极高的成本。不同于大部分需要收集机器人数据进行学习的方法，本方法利用生成式大模型stable diffusion蕴含的语义信息，通过从人类视频演示中提取精确的抓取affordance，学习物体的抓取语义，极大地减少了需要灵巧手抓取所需要的成本，并能够轻松泛化到各种灵巧手上，最终仅依赖单个深度相机，就可以达到90%的抓取正确率，并能轻松泛化到同类物体和不同位置

\begin{abstract}

Dexterous grasp synthesis must jointly satisfy functional intent and physical feasibility, yet existing pipelines often decouple semantic grounding from refinement, yielding unstable or non-functional contacts under object and pose variations. This challenge is exacerbated by the high dimensionality and kinematic diversity of multi-fingered hands, which makes many methods rely on large, hardware-specific grasp datasets collected in simulation or through costly real-world trials. We propose a data-efficient framework that bypasses robot grasp data collection by exploiting object-centric semantic priors in pretrained generative diffusion models. Temporally aligned and fine-grained grasp affordances are extracted from raw human video demonstrations and fused with 3D scene geometry from depth images to infer semantically grounded contact targets. We further incorporate these affordance regions into the grasp refinement objective, explicitly guiding each fingertip toward its predicted region during optimization. 
%The resulting system produces stable, functionally appropriate multi-contact grasps across common objects and tools, while exhibiting strong generalization across previously unseen object instances within a category, pose variations, and multiple hand embodiments.
The resulting system produces stable, human-intuitive multi-contact grasps across common objects and tools, while exhibiting strong generalization to previously unseen object instances within a category, pose variations, and multiple hand embodiments.
This work (i) introduces a semantic affordance extraction pipeline leveraging vision--language generative priors for dexterous grasping, (ii) demonstrates cross-hand generalization without constructing hardware-specific grasp datasets, and (iii) establishes that a single depth modality suffices for high-performance grasp synthesis when coupled with foundation-model semantics.
Our results highlight a path toward scalable, hardware-agnostic dexterous manipulation driven by human demonstrations and pretrained generative models.
\end{abstract}

%%%%%%%%%%%%%%%%%%%%%%%%%%%%%%%%%%%%%%%%%%%%%%%%%%%%%%%%%%%%%%%%%%%%%%%%%%%%%%%%
\section{Introduction}

Reliable grasp synthesis remains a central, yet unresolved, challenge in dexterous robot manipulation. Compared with parallel-jaw grippers, multi-fingered dexterous hands offer substantially higher kinematic expressivity and potential contact richness, but this increase in dimensionality and hand--object--environment interaction complexity dramatically amplifies the difficulty of learning and deploying robust grasp strategies. The classical problem can be informally factorized into (i) ``\emph{how to grasp}": the structured, role-aware coordination of fingers, contacts, approach direction, and hand synergies, and (ii) ``\emph{where to grasp}": the spatial localization of functionally exploitable object regions. Despite rapid progress, existing methods typically address these two issues in isolation, leading to brittle policies, embodiment entanglement, and limited cross-object generalization.

On ``\emph{how} to grasp", recent works leverage reinforcement learning or large-scale supervised training in simulation to explore expansive action spaces~\cite{xu2023unidexgrasp,wang2022dexgraspnet}. However, simulation-trained policies customarily rely on privileged, complete, and noise-free geometric states, which are rarely attainable in real-world deployments. This induces significant sim-to-real gaps when facing realistic sensing artifacts such as partial views, self-occlusions, depth discontinuities, and modality noise.
An alternative line---direct large-scale real-world data collection (e.g., AnyDexGrasp~\cite{fang2025anydexgrasp})---suffers from prohibitive operational cost and still struggles to generalize to unseen dexterous hand embodiments. A core culprit is representational overfitting: action/policy parameterizations are tightly coupled to a specific hand model's joint layout, contact affordances, and actuation limits, precluding systematic transfer.

On ``\emph{where} to grasp", prevailing affordance research yields coarse region-of-interest predictions~\cite{luo2023learning,luo2022learning}. While such representations offer a spatial prior for potential contact regions, they stop short of encoding the per-finger engagement strategy and the geometry-conditioned sequencing necessary for stable multi-contact closure. 
Thus, although they identify candidate regions, they fail to specify the fine-grained per-finger instructions required for dexterous manipulation (e.g., contact assignments, approach vectors, and local-geometry adaptations). Recasting contact reasoning as keypoint detection is likewise insufficient: reliable contact loci often lack discriminative RGB cues and hinge on object-part semantics and inter-finger relations that appearance-only backbones (e.g., CNNs/ViTs~\cite{dosovitskiy2021an}) do not capture.

Our core idea to tackle the above challenges is that Internet-scale text-to-image diffusion models internalize multi-level knowledge about objects, parts, materials, and functional geometry ~\cite{rombach2022high}. 
Prior work~\cite{yang2023boosting} shows that intermediate denoising features encode exactly the semantic priors missing in standard discriminative backbones, and that these features can be repurposed—not to generate data, but to ground grasp semantics.

Building on this insight, we infer a \emph{Finger-Specific Affordance Field (FSAF)} that unifies the ``where'' and ``how'' of dexterous contacts: an object-conditioned field that assigns each surface location a finger-conditioned likelihood and role descriptor, capturing feasible contact placement and inter-finger compatibility.
We repurpose a frozen Stable Diffusion~\cite{rombach2022high} U\hbox{-}Net as a semantic backbone to predict FSAF from image--text inputs, and integrate it into an optimization-based grasp refinement and execution pipeline.
Given the inferred FSAF, we synthesize finger-consistent grasps without additional teleoperation by warm-starting refinement toward the predicted regions, and transfer across heterogeneous dexterous hands by changing only the kinematic model (no retraining).
Experiments show strong generalization to seen and unseen objects, improved affordance grounding, and high real-robot grasp success with cross-embodiment transfer.

In summary, we make the following contributions:
\begin{itemize}
    \item \textbf{Finger-Specific Affordance Field (FSAF).} We introduce a fine-grained, per-finger affordance representation that leverages vision--language generative priors to jointly encode object affordance functions and contact-level manipulation semantics, decomposing monolithic grasp poses into semantically grounded finger role assignments.
    \item \textbf{Demonstration-light, affordance-conditioned grasp synthesis.} Our grasp generation consumes the inferred FSAF to synthesize stable, role-constrained multi-finger grasp configurations without requiring large-scale teleoperation demonstration corpora, substantially reducing data collection burden.
    \item \textbf{Cross-embodiment generalization.} We achieve direct transfer of grasp strategies across heterogeneous dexterous hands, evidencing that the proposed affordance abstraction disentangles grasp semantics from specific embodiment kinematics.
\end{itemize}

\section{RELATED WORK}

\subsection{Learning from Human Demonstration Videos}

Learning from human demonstration videos is central to robot learning because such data are easier to collect and encode strong task priors~\cite{luo2022learning,ren2025motion}; however, a persistent embodiment gap separates human hands from robotic dexterous end-effectors. The prevailing remedy is retargeting—mapping human motions to robot kinematics~\cite{zhao2024dexh2r,li2025maniptrans}—which is effective for teleoperation and data gathering but typically still requires reinforcement learning in simulation to attain dexterous proficiency, thereby introducing a sim-to-real gap and limiting cross-embodiment generalization. Recently, a more fundamental route is to learn manipulation structure from human hand data that robots can directly consume. For parallel-jaw grippers, PointPolicy~\cite{haldar2025point} uses video-derived keypoints to supervise gripper control; however, such position-only cues under-specify dexterous-hand interaction. In this work, we therefore learn finger-specific affordances from human demonstrations—object-conditioned descriptors that specify, per finger, contactable regions to represent how dexterous hands should interact with objects.
\begin{figure*}[t]
\centering
\includegraphics[width=\textwidth]{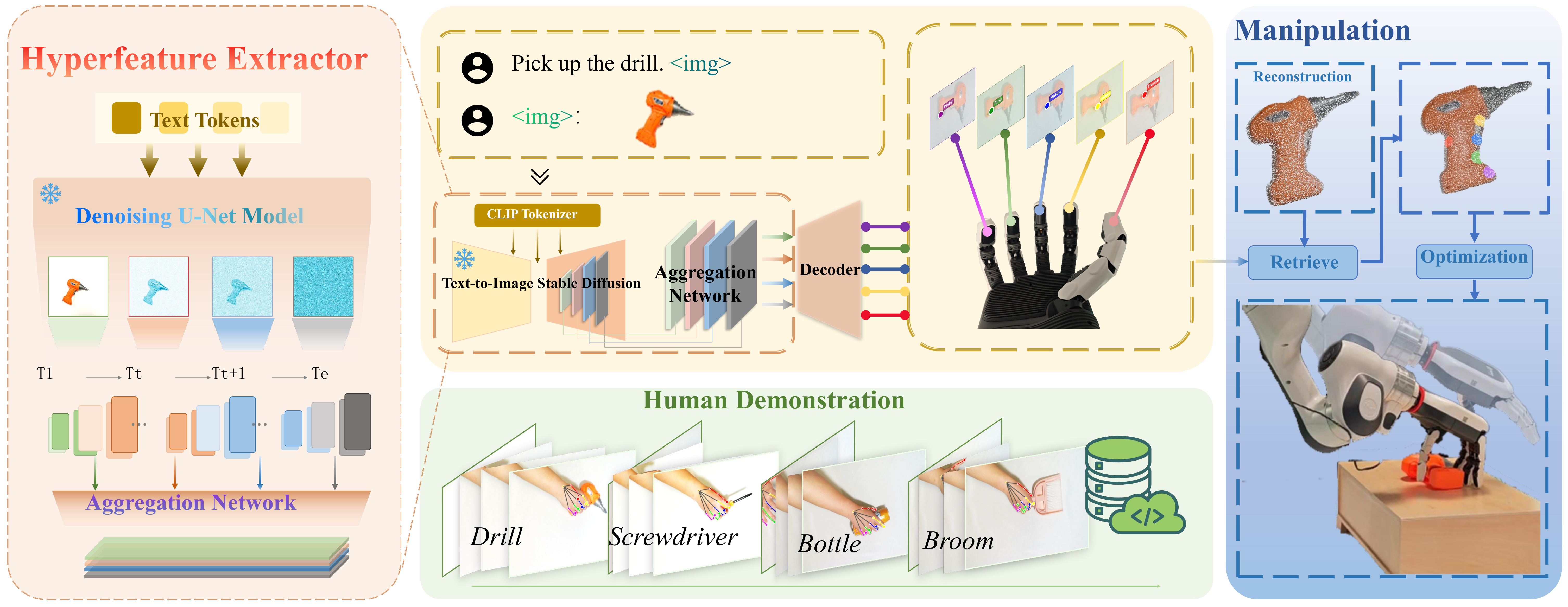}
\caption{\textbf{Pipeline overview.}
(1) \emph{Hyperfeature extraction:} A frozen text-to-image diffusion U-Net encodes the object image with text conditioning. Multi-timestep, multi-scale activations are aggregated into hyperfeatures $A_g$.
(2) \emph{Finger-specific affordance grounding:} An FPN-style decoder maps $A_g$ to five per-finger likelihood maps $\hat{H}$ supervised by fingertip labels from human demonstrations.
(3) \emph{Manipulation:} We reconstruct the object with SAM3+SAM3D, lift peaks of $\hat{H}$ to 3D contacts, and augment GraspQP with a finger-specific affordance term to refine stable, human-intuitive grasps.}
\label{fig:pipeline}
\vspace{-10pt}
\end{figure*}
\subsection{Object-centric Representation Learning}
\label{subsec:Object-centric Representation Learning}

Much of object-centric grasping is tailored to parallel-jaw grippers, representing grasps by 2D/3D locations. Such position-only cues under-specify dexterous-hand interactions \cite{srirama2024hrp,dong2025rtagrasp}. In addition, a subset of dexterous hand approaches relies on template matching, built around exemplars specific to the object or the task, thus having limited generalizability to novel objects/tasks~\cite{kokic2020learning}. Contact-centric representations improve task relevance but require complex pipelines and costly data collection to obtain dexterous-hand contact regions~\cite{yang2022oakink}. More recently, interaction-driven approaches such as CMKA~\cite{yang2025multi} learn interaction regions from web imagery, yet map object keypoints to wrist, index, and little-finger anchors, effectively producing a single grasp parameterized by three points and limiting performance to on-shelf settings. These limitations motivate per-finger contact representations rather than region- or point-only cues.
We learn per-finger contact cues and, via depth-and-segmentation projection, map them onto object surfaces, to further derive precise tabletop grasps.

\subsection{Affordance Learning}
% Affordance is crucial for robotic grasping and manipulation in dynamic, physical environments. Previous research has achieved remarkable progress in learning affordance knowledge from Human-Object-Interaction (HOI) images~\cite{shao2024great} and videos~\cite{bahl2023affordances}, language instructions~\cite{li2024laso}, and 3D point clouds~\cite{delitzas2024scenefun3d}. More recently, researchers have expanded the reasoning ability of multimodal large language models into affordance learning~\cite{wang2025affordance,qian2024affordancellm}, achieving advanced affordance reasoning and generalization abilities. Moreover, the text-to-image diffusion model~\cite{rombach2022high} shows promising affordance-aware understanding ability by generating precise hand-object images~\cite{yang2023boosting}. 
However, most approaches localize coarse object regions (e.g., boxes or masks), which is insufficient for fine-grained, finger-specific guidance required by dexterous hands—motivating representations at the per-finger contact level.
% Affordance learning has been studied across multiple modalities, including HOI images/videos~\cite{yang2023grounding,shao2024great,bahl2023affordances,luo2023learning} and 3D point clouds~\cite{deng20213d,delitzas2024scenefun3d}, with recent extensions leveraging multimodal reasoning models~\cite{wang2025affordance,qian2024affordancellm}. Text-to-image diffusion models further exhibit strong interaction priors by synthesizing realistic hand--object configurations~\cite{rombach2022high,yang2023boosting}. 
% However, most existing methods provide coarse region-level cues (e.g., boxes or masks) rather than finger-conditioned contact targets, which is insufficient for dexterous hands and motivates per-finger contact-level representations.

\section{METHODS}

We present a perception-to-optimization pipeline that learns \emph{finger-specific affordances} from human demonstrations and uses them to synthesize executable dexterous grasps (Fig.~\ref{fig:pipeline}). 
We extract semantically grounded hyperfeatures using a frozen text-to-image diffusion model~\cite{rombach2022high} and decode them into five per-finger likelihood maps, which provide finger-wise contact priors on the object surface. 
We then formulate grasp synthesis as an \emph{affordance-conditioned} optimization: the per-finger priors are incorporated as a finger-wise alignment term alongside force-closure and feasibility objectives, thereby coupling grasp pose generation with the learned affordance priors.

\subsection{Data Collection}
We construct a finger-wise contact affordance dataset from human demonstration videos using a custom data acquisition pipeline. For each recorded grasp sequence, we first automatically detect 2D hand keypoints with the RTMPose~\cite{jiang2023rtmpose} hand pose estimator. We then identify (i) an \emph{object-only} keyframe in which the target object is fully visible and no hand pixels are present, and (ii) the earliest \emph{stable grasp} frame in which the human hand establishes clear contact with the object. This pair of frames form one training sample: the object-only frame provides a clean visual context of the object's geometry and texture, while the grasp frame provides the supervisory signal for finger-specific contact localization.

On the image lattice $\Omega=\{0,\ldots,h-1\}\times\{0,\ldots,w-1\}$, each fingertip $k$ with center $\mu_k$ induces a Gaussian channel
$H_k(u)=\exp(-\|u-\mu_k\|_2^2/(2\sigma^2))$ for $u\in\Omega$; stacking yields a tensor $H$ of shape $5\times h\times w$.
We set $\sigma=\min(h,w)/64$ and train the predictor $\hat H$ to regress $H$ using MSE.

\subsection{Finger-Specific Affordance Representation from Diffusion Models}
\label{subsec:finger_affordance}

\subsubsection{Feature synthesis and hyperfeature aggregation}
Empirical evidence that diffusion models generate precise hand–object grasp~\cite{yang2023boosting} scenarios indicates that Stable Diffusion~\cite{rombach2022high} has already acquired the grasp-critical feature space, integrating holistic affordance semantics with nuanced image details via its generative learning process. Inspired by prior work~\cite{yang2023boosting,luo2023diffusion}, we leverage the rich semantic grounding capability of large pretrained text-to-image diffusion models to localize finger-specific contact affordances on object images. Distinct from previous approaches, we jointly harvest both visual and textual affordance semantics from Stable Diffusion. 
Given an image--text pair $(x_0, s)$, we encode a latent $z_0=E(x_0)$ with the variational autoencoder (VAE)
and follow the standard \emph{latent} forward noising process
\begin{equation}
z_t \;=\; \sqrt{\bar{\alpha}_t}\,z_0 \;+\; \sqrt{1-\bar{\alpha}_t}\,\epsilon, 
\qquad \epsilon \sim \mathcal{N}(0, I),
\label{eq:forward}
\end{equation}
where $\bar{\alpha}_t=\prod_{\tau=1}^{t}\alpha_\tau$ as defined in ~\cite{rombach2022high}.
We keep the text encoder $T(\cdot)$ frozen and obtain token embeddings $c=T(s)$.
For computational reasons, we run the diffusion network for $T$ timesteps but only select
a subset $\mathcal{S}$ with $|\mathcal{S}|=S$ for feature aggregation.
Feeding $(z_t, t, c)$ into a frozen diffusion U\hbox{-}Net yields multi-scale activations:
\[
\big\{A^{(t)}_{v,1},\,A^{(t)}_{v,2},\,\ldots,\,A^{(t)}_{v,L}\big\},
\quad t\in\mathcal{S},
\]
Here, $A^{(t)}_{v,\ell}\in\mathbb{R}^{B\times C_\ell\times H_\ell\times W_\ell}$ denotes the feature map from the $\ell$-th U\hbox{-}Net block at timestep $t$.

To summarize complementary semantics across scales and timesteps, we attach lightweight
bottlenecks $\{b_\ell\}_{\ell=1}^L$ (shared across timesteps) and learn mixing coefficients
$\{w_{\ell,t}\}_{\ell=1,t\in\mathcal{S}}^{L}$ to produce a global affordance descriptor:
\begin{equation}
A_g \;=\; \sum_{t\in\mathcal{S}}\;\sum_{\ell=1}^{L} w_{\ell,t}\;\, b_\ell\!\left(A^{(t)}_{v,\ell}\right).
\label{eq:agg}
\end{equation}
In practice, each $b_\ell$ is a $1{\times}1$ convolution followed by global average pooling
to a fixed $d$-dimensional vector.
We parameterize $\tilde{w}_{\ell,t}$ freely and apply a softmax over all $(\ell,t)$.

\subsubsection{Finger-Specific Affordance Heatmap Prediction}

\begin{figure}[t]
\centering
\includegraphics[width=\linewidth]{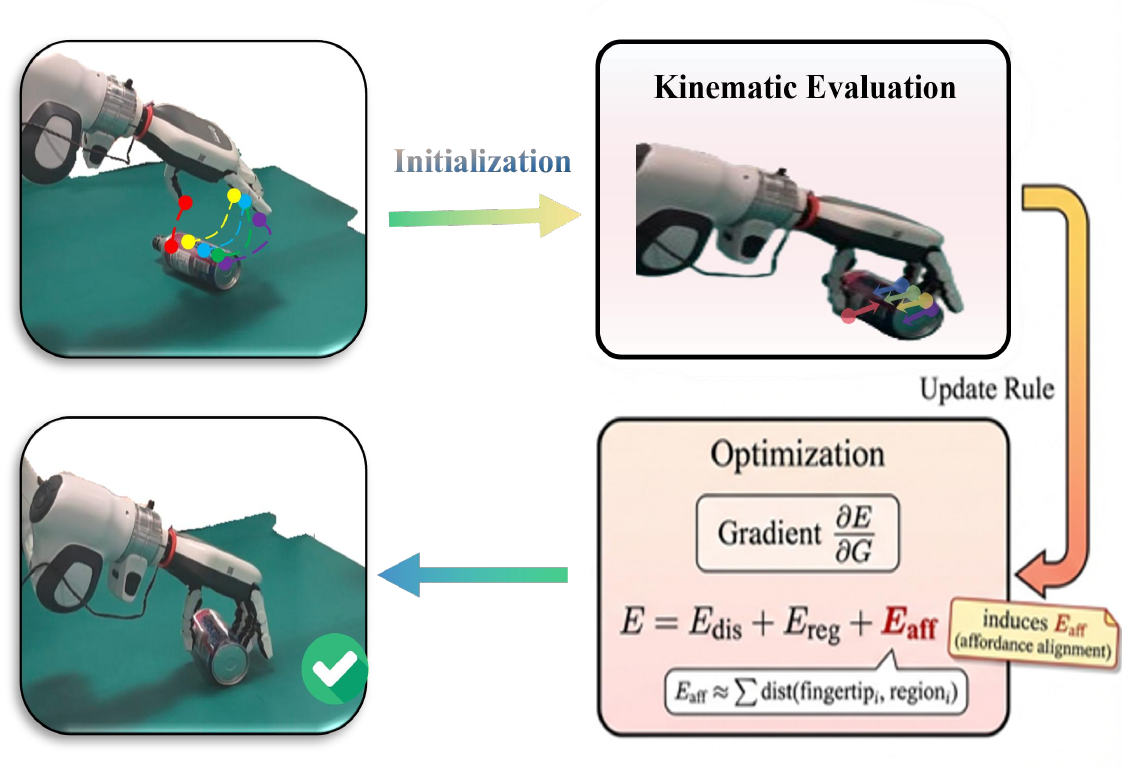}
\caption{ Finger-Specific Affordance Field (FSAF) regions on the object induce an affordance-alignment term $E_{\mathrm{aff}}$.}
\label{fig:motion_pipeline}
\vspace{-12pt}
\end{figure}

To enhance the fine-grained finger-specific affordance grounding ability while preserving high-level semantic discrimination and recovering spatial details, we proposed a feature pyramid network (FPN)–style decoder, which progressively upsamples $A_g$ while injecting lateral features from shallower resolutions, producing a dense fused feature map suitable for contact prediction.

% The FPN decoder takes global attention features $A_g \in \mathbb{R}^{B \times N \times H \times W}$ and produces multi-scale segmentation outputs through lateral connections and top-down fusion:
% \begin{equation}
% \hat{H}_k(u) = \text{Decoder}(A_g),
% \end{equation}
% where the decoder performs hierarchical feature extraction at multiple scales $\{256, 128, 64\}$ channels, followed by progressive upsampling and feature fusion to generate the final 5-channel segmentation heatmap $\hat{H}_k(u) \in \mathbb{R}^{B \times 5 \times H_{out} \times W_{out}}$.

Given the dense hyperfeature $A_g\!\in\!\mathbb{R}^{B\times C\times h\times w}$, we decode it with
three lateral $1{\times}1$ projections $\{\phi_r\}_{r=3}^{1}$ (to channels $c_r\!\in\!\{256,128,64\}$)
and a top–down pathway. Let target sizes $s_3{=}(h,w)$, $s_2{=}(2h,2w)$, $s_1{=}(H_{\!out},W_{\!out})$,
$U_s$ be bilinear upsampling to size $s$, $\tau_r$ a $1{\times}1$ adapter, and $\psi_r$ a
$3{\times}3$ smoothing conv with nonlinearity. With $F_{4}\!\equiv\!0$, we compute
\begin{equation}
F_r \;=\; \psi_r\!\Big( U_{s_r}\big(\tau_r(F_{r+1})\big) \;+\; U_{s_r}\big(\phi_r(A_g)\big) \Big),
\quad r=3,2,1,
\label{eq:fpn-rec}
\end{equation}
and obtain the per-finger heatmaps by a final $3{\times}3$ projection
\begin{equation}
\hat H \;=\; \kappa(F_1)\in\mathbb{R}^{B\times K\times H_{\!out}\times W_{\!out}},\quad K=5.
\label{eq:fpn-out}
\end{equation}

% Given $A_g\!\in\!\mathbb{R}^{B\times C\times h\times w}$, an FPN-style decoder produces per-finger heatmaps
% \[
% \hat H \;=\; \mathrm{Dec}(A_g)\;\in\;\mathbb{R}^{B\times 5\times H_{\mathrm{out}}\times W_{\mathrm{out}}},
% \]
% where $\mathrm{Dec}$ applies three $1{\times}1$ lateral projections (to 256/128/64 channels) and a top–down fusion pathway with progressive upsampling and $3{\times}3$ smoothing. Elementwise addition at matched resolutions is used for lateral–topdown fusion. For notation, $\hat H_k(u)$ denotes the $k$-th channel value of $\hat H$ at pixel $u$.

We optimize a mean squared error (MSE) loss over all fingers and pixels:
\begin{equation}
\mathcal{L}_{\mathrm{MSE}} = \frac{1}{5|\Omega|}\sum_{k=1}^{5} \sum_{u \in \Omega} \bigl(\hat H_k(u) - H_k(u)\bigr)^2.
\label{eq:mse_loss}
\end{equation}

\subsection{Finger-Specific Affordance Matching for Dexterous Grasping}

\textbf{Object Reconstruction and Affordance Grounding.}  
% Our perception pipeline begins by segmenting the target object using SAM3~\cite{carion2025sam3segmentconcepts}. The segmentation mask is combined with dense depth data from an RGB-D sensor, and the corresponding pixels are back-projected into 3D coordinates via the camera's intrinsic parameters to form a scene point cloud. This point cloud is subsequently processed by SAM3D~\cite{sam3dteam2025sam3d3dfyimages} to reconstruct the 3D object mesh and estimate its 6-DoF pose (rotation, translation, and scale). This initial pose is then refined through ICP registration against the scene point cloud to achieve a precise object-to-camera transformation $\mathbf{T}_{\mathrm{obj} \to \mathrm{cam}} \in SE(3)$.

We first segment the target object with SAM3~\cite{carion2025sam3segmentconcepts} and back-project the masked RGB-D depth into a partial object point cloud using the camera intrinsics. We then apply a reconstruction pipeline based on SAM3D~\cite{sam3dteam2025sam3d3dfyimages} to obtain a posed object surface, and refine the estimated pose via ICP. The resulting object-to-camera transform is denoted as $\mathbf{T}_{\mathrm{obj}\to \mathrm{cam}} \in SE(3)$.
% \textbf{Affordance Point Projection via Raycasting.}
% To localize each predicted finger keypoint on the reconstructed object surface, we perform raycasting against the posed mesh in the camera coordinate frame. For each 2D keypoint $(u_k, v_k)$ in the image, we construct a ray from the camera origin along the direction $\mathbf{d}_k = \mathbf{K}^{-1}[u_k, v_k, 1]^\top$, where $\mathbf{K}$ denotes the camera intrinsic matrix. The ray--mesh intersection yields a 3D contact point $c_k^{\mathrm{cam}}$ and the triangle face normal $\hat{n}_k^{\mathrm{cam}}$ at the hit location. For keypoints whose rays do not directly intersect the mesh (e.g., due to minor calibration errors or partial occlusion), we employ a closest-point fallback: we query the nearest mesh surface point at an approximate depth and retrieve its vertex normal, ensuring that all predicted keypoints are robustly anchored to the object geometry. All resulting contact points and normals are subsequently transformed to the robot base frame via hand--eye calibration.

Using the finger-specific affordances derived in Section~\ref{subsec:finger_affordance}, we re-project the object’s point cloud onto the image plane. For each predicted finger keypoint, we identify the nearest N point-cloud projections on the image, forming a set of robust fingertip contact candidates.

\textbf{Grasp Refinement and Integration of Finger-Specific Affordances.}  
Our grasp refinement module is inspired by GraspQP~\cite{graspqp2025} and we introduce key innovations by refining this approach to account for finger-specific affordance cues, allowing for more precise and targeted grasp synthesis.

We minimize the following energy function to ensure the grasp is physically stable and meanwhile, aligned with the affordance regions of each finger:
\begin{equation}
E = E_{\mathrm{fc}} + w_{\mathrm{dis}} E_{\mathrm{dis}} + w_{\mathrm{reg}} E_{\mathrm{reg}} + w_{\mathrm{aff}} E_{\mathrm{aff}},
\end{equation}
where \(E_{\mathrm{fc}}\) is a force-closure term from GraspQP that optimizes contact force distributions to balance external wrenches under Coulomb friction. The term \(E_{\mathrm{dis}}\) penalizes the distance between active contact points \(\mathcal{C}'\) and the object surface, ensuring proper contact placement, while \(E_{\mathrm{reg}}\) consists of regularization terms designed to prevent hand-object penetration, joint-limit violations, and self-penetration.

We incorporate Finger-Specific Affordance Fields (FSAF) into grasp refinement as a finger-conditioned geometric prior. 
\textbf{Affordance prior in the objective.} Concretely, to instantiate the affordance regularizer in the refinement objective defined above, we add the following contact-to-region alignment term:
\begin{equation}
E_{\mathrm{aff}}=\frac{1}{|\mathcal{C}|}\sum_{j\in\mathcal{C}}\min_{q\in S_{g(j)}}\|c_j-q\|_2^2,
\end{equation}
where $\mathcal{C}$ denotes the set of contact points optimized during refinement, and $g(j)$ maps each contact $c_j$ to its corresponding finger-specific affordance region $S_{g(j)}$. Minimizing the full objective thus encourages contacts to stay on their predicted regions while the remaining terms enforce physical feasibility.

\textbf{Initialization (warm start).} We warm-start the refinement solver with an affordance-consistent contact layout: for each finger $i$, we initialize its fingertip contact $c_i$ near $S_i$ by enforcing $\mathrm{dist}(c_i,S_i)\le \delta$ with $\delta=3\,\mathrm{cm}$. This narrows the search to role-consistent configurations and reduces sensitivity to local minima.

\textbf{Execution.} After optimizing the objective to obtain the refined grasp posture, we execute a three-phase motion plan---\emph{approach}, \emph{close}, and \emph{hold}---to realize smooth finger trajectories and stable contact establishment. We evaluate grasp quality and task performance in Sec.~\ref{app:func_affordance_grasp}.

\section{EXPERIMENTS}

In this section, we conduct comprehensive experiments to validate our method from three complementary perspectives:
\textbf{RQ1:} Does the proposed finger-specific grasp representation outperform prior keypoint/affordance grounding, and does leveraging vision–language priors from Stable Diffusion model yield gains over discriminative backbones?\\
% \textbf{RQ2:} Does the method improve the functional correctness and stability of dexterous grasps on seen and unseen objects, even for objects with weak part cues?\\
\textbf{RQ2:} Does the method improve grasp stability and produce more human-intuitive contact placements on seen and unseen objects, especially for categories with weak part cues?\\
\textbf{RQ3:} Does the learned finger-specific affordance representation transfer across dexterous hands with minimal adaptation?

\subsection{Experimental Setup}

\begin{figure}[t]
  \centering
  \includegraphics[width=0.8\linewidth]{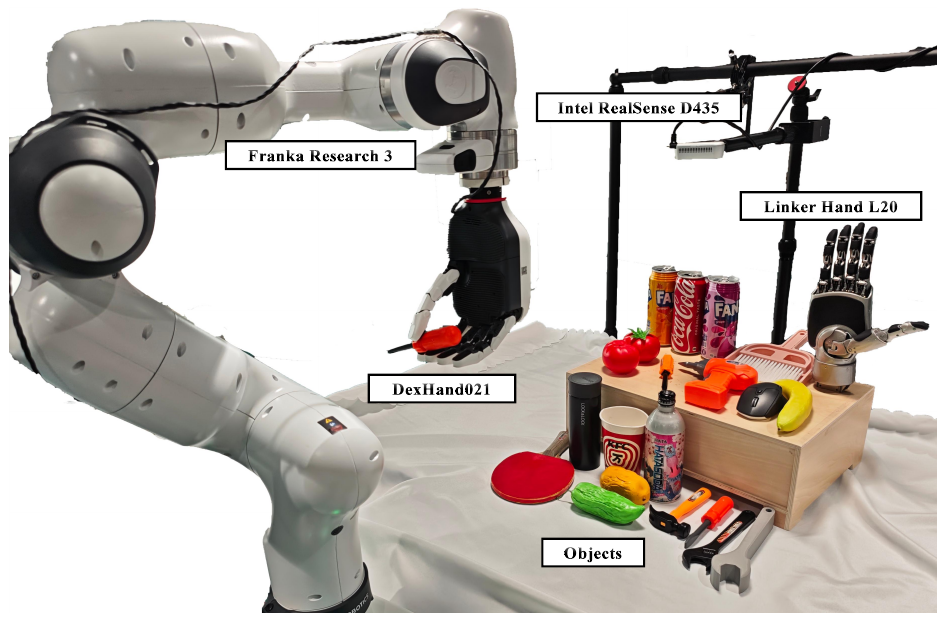} % you can change the filename freely
  \caption{Robotic arm and dexterous-hand experimental platform used in our study.}
  \label{fig:exp_platform}
  \vspace{-12pt}
\end{figure}

\textbf{Dataset:} We collect 130 \emph{finger-specific} human grasp demonstrations over 13 everyday objects (10 per object; e.g., banana, screwdriver, bottle). For evaluation, we test on seven \emph{unseen} objects that include both (i) new instances of previously seen categories and (ii) entirely novel categories (e.g., wrench, hammer).

\begin{figure*}[t]
    \centering
    \includegraphics[width=0.95\linewidth]{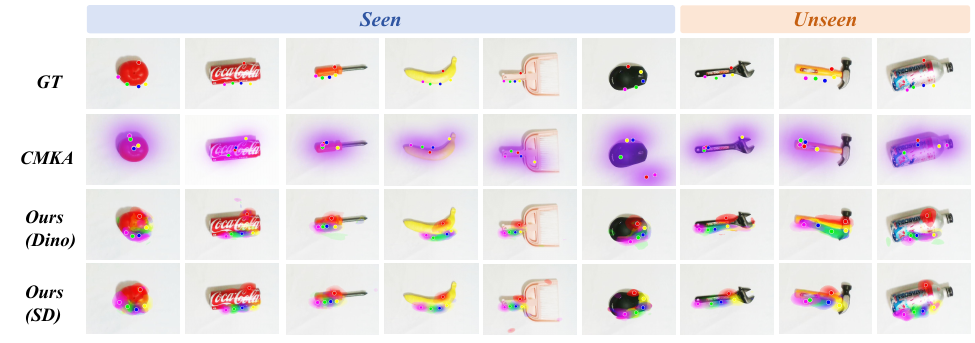}
    \caption{\textbf{Qualitative comparison of finger-specific affordance grounding.}
    Left: \emph{Seen} objects; Right: \emph{Unseen} objects. Rows show CMKA, Ours (DINO features), and Ours (Stable Diffusion). Overlays visualize five per-finger likelihood maps; colored dots indicate annotated fingertip contacts. The diffusion-based variant produces sharper, finger-disentangled hotspots aligned with functional parts and preserves localization quality on unseen tools. CMKA produces keypoints directly and a single-channel heatmap is generated via post-processing following CMKA, thus its overlays are rendered in a single purple color, whereas our method outputs five-channel heatmaps rendered with five distinct colors.}
    \label{fig:compare}
    \vspace{-12pt}
\end{figure*}

\textbf{Implementation Details:} We train for $4{,}000$ steps with batch size $2$ on a single NVIDIA H100 GPU using AdamW (initial learning rate $10^{-3}$ with cosine decay). Raw RGB frames are resized to $640\times320$ with letterboxing when necessary to preserve aspect ratio, and pixel intensities are normalized to $[0,1]$. Predicted and ground-truth finger heatmaps are uniformly rescaled to $448\times448$ following saliency-grounding conventions, and each per-finger heatmap is re-normalized to integrate to $1$.
% to prevent trivial magnitude scaling. 
All results are averaged over 3 random seeds.
The robotic arm--dexterous-hand platform is shown in Fig.~\ref{fig:exp_platform}; the real-world setup comprises two dexterous hands (DexHand021 and Linker Hand L20), a Franka arm, and a RealSense~435 camera.

\textbf{Evaluation Metrics:}
For finger-specific affordance grounding, We report KLD, SIM, and NSS~\cite{bylinskii2018different}, \textbf{following CMKA~\cite{yang2025multi} for metric computation}:
\begin{itemize}
\item \textbf{KLD} (lower is better): Kullback--Leibler divergence between predicted and ground-truth heatmaps.
\item \textbf{SIM} (higher is better): Normalized histogram intersection (similarity).
\item \textbf{NSS} (higher is better): Normalized Scanpath Saliency; per finger we sample the annotated keypoint location as a fixation.
\end{itemize}
For Dexterous grasping, we follow the definitions of metrics provided
 in [4] and [5]: 
\textbf {Grasping Success Rate (Suc. R.)}: A grasp is considered
 successful if the object can be lifted above 0.1 m and
 held stable longer than 3 seconds. Here, grasping specifically refers to grasping based on manipulation affordances.

\subsection{Results of Finger-Specific Affordance Grounding}
\label{subsec:finger_specific_affordance}
\begin{figure}[t]
    \centering
    \includegraphics[width=0.95\linewidth]{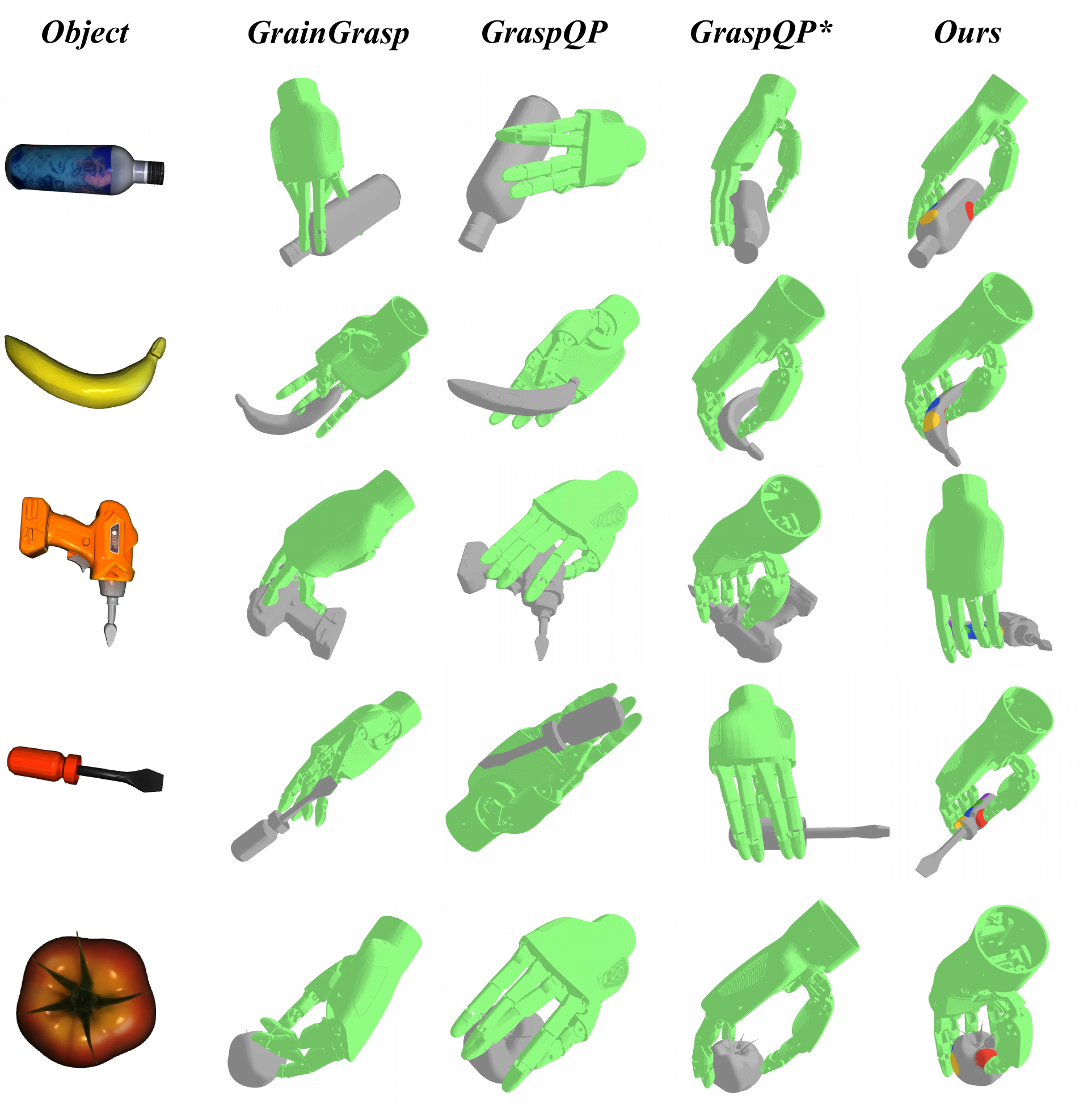}
    \caption{\textbf{Qualitative grasp visualization.} We visualize predicted grasps on five objects (bottle, banana, drill, screwdriver, and tomato). From left to right, we compare GrainGrasp, GraspQP with random initialization, GraspQP* (GraspQP initialized in the same way as ours), and our method. In the last column, the colored surface regions on the object mesh indicate our predicted finger-specific affordances.}
    \label{fig:grasp_compare}
     \vspace{-12pt}
\end{figure}
Because there is no prior work that directly couples vision-language models with fine-grained, finger-specific affordance keypoints, we select the closest approach as the baseline. 
CMKA~\cite{yang2025multi} leverages exocentric (third-person) interaction images as input, applies SAM~\cite{kirillov2023segment} for multi-scale object segmentation, and then performs clustering over the segmentation-driven feature hierarchy to learn object affordance keypoints.
To ensure fairness, we train CMKA~\cite{yang2025multi} using the exocentric interaction images from \textbf{our dataset}, keeping all other hyperparameters unchanged. For metric computation, we follow a procedure consistent with CMKA: the predicted keypoints are first rescaled to the original image resolution, then converted into heatmaps by placing Gaussian kernels centered at each keypoint. To validate the necessity of large-scale generative models in our framework, We ablate feature extractors by replacing SD hyperfeatures with CLIP~\cite{radford2021learning} and DINO~\cite{caron2021emerging}
% we conduct ablations that replace the diffusion\mbox{-}derived hyperfeatures with discriminative backbones CLIP~\cite{radford2021learning} and DINO~\cite{caron2021emerging} used purely as feature extractors.
% The original CMKA formulation differs from ours in two important ways: (i) it predicts three contact points corresponding to the index finger, little finger, and the wrist projection region, rather than all five fingers; (ii) it is trained on the FAH dataset, which only contains tool-like objects, limiting generalization to everyday categories.

% \newcolumntype{L}{>{\centering\arraybackslash}p{1.5cm}} % 

\begin{table}[t]
\centering
\caption{Comparison and ablation on the finger-specific affordance grounding benchmark. Best results in \textbf{bold}. ($\uparrow/\downarrow$ denote higher/lower is better.)}
\label{tab:comparison_ablation}
\renewcommand{\arraystretch}{1.25}
\setlength{\tabcolsep}{6pt}
\small
\begin{tabular}{lccc}
\toprule
Model / Variant & KLD($\downarrow$) & SIM($\uparrow$) & NSS($\uparrow$) \\
\midrule
CMKA~\cite{yang2025multi}             & 11.184 & 0.177 & 1.861 \\
\midrule
Ours (CLIP)      & 6.690  & 0.355 & 3.815 \\
Ours (DINO)      & 3.301  & 0.473 & 5.016 \\
\midrule
Ours (SD)      & \textbf{2.491} & \textbf{0.551} & \textbf{5.518} \\
\bottomrule
\end{tabular}
\vspace{-10pt}
\end{table}

Table~\ref{tab:comparison_ablation} shows that CMKA underperforms on finger-specific affordance grounding: its segmentation-driven hierarchy is brittle when objects lack distinctive, part-aligned visual regions. Fig.~\ref{fig:compare} (seen vs. unseen) illustrates typical failures, where predictions collapse toward object centroids or spill across boundaries, yielding contacts that miss functional parts. In contrast, our method improves all three metrics (lower KLD, higher SIM/NSS), indicating sharper, part-aligned localization rather than generic centers or silhouette edges.

Holding data, labels, losses, and the training schedule fixed while varying only the feature extractor, we isolate the contribution of diffusion hyperfeatures. Compared with DINO/CLIP, \textbf{KLD} increases by \(1.33\times\)/\(2.69\times\), while \textbf{SIM} drops by \(14.2\%\)/\(35.6\%\) and \textbf{NSS} by \(9.1\%\)/\(30.9\%\). Qualitatively, DINO features more often merge adjacent fingers and produce broader activations with boundary spill-over; on unseen elongated tools, they also bias toward visually salient parts rather than human-intuitive grasp regions. Diffusion hyperfeatures consistently suppress three dominant errors—centroid collapse, boundary spill-over, and fingertip swapping—supporting \textbf{RQ1} that they are critical for precise per-finger affordance localization.

\subsection{Functional Affordance Grasping with High-DOF Hands}
\label{app:func_affordance_grasp}

Our evaluation targets \emph{on-table}, affordance-driven dexterous grasping with finger-specific contacts and human-consistent grasp sites.
Existing simulation-trained policies~\cite{xu2023unidexgrasp} optimize for success in simplified settings and cannot select grasp sites following human conventions (e.g., grasping a hammer by its head).
To assess alignment with human grasp choices within our setting, we include two point-cloud-based imitation learning policies (Diffusion Policy 3D~\cite{ze20243d} and ACT-3D~\cite{gervet2023actd}) trained from our teleoperated demonstrations (30 trajectories).
We also compare to CMKA, representing a recent affordance-based dexterous grasping approach. We evaluate CMKA strictly under its original, unmodified pipeline: we directly use the three keypoints predicted by CMKA, lift them to 3D, assign them to the functional finger, little finger, and wrist, and then execute the method’s fixed grasp execution strategy.
We additionally compare with GrainGrasp~\cite{zhao2024graingrasp}, a representative fine-grained contact-guided grasp synthesis method, by adapting its contact-guided refinement stage to our robotic hand embodiment and evaluating it under the same perception and execution pipeline for real-robot success-rate comparison.

Fig.~\ref{fig:grasp_compare} presents qualitative grasp predictions on the Linker Hand L20 for GrainGrasp, GraspQP, and our method. Our approach consistently produces human-intuitive grasp poses: guided by finger-specific affordances, the fingertips align tightly with the predicted graspable regions, yielding contact layouts that match the functional geometry of each object. 
In contrast, GraspQP with random initialization performs markedly worse, often converging to non-human-like poses (e.g., on the \emph{screwdriver}), missing the intended grasp regions, and occasionally producing degenerate configurations with overlapping fingers. 
GraspQP* (GraspQP with the same region-conditioned initialization as our method) can sometimes achieve similarly reasonable grasps on \emph{bottle}, \emph{tomato}, and \emph{screwdriver}, indicating that a strong warm start already alleviates some local-minima failures; however, it still fails on the \emph{banana} and converges to visually implausible configurations on the \emph{drill}, suggesting that initialization alone is insufficient to enforce affordance-consistent, finger-wise contact placement. 
It's also noteworthy that GrainGrasp fails to generate meaningful grasps across these objects, as it is tailored to MANO-scale human hands and its contact representation and kinematic assumptions do not transfer well to high-DOF robotic hands.
% Finally, GrainGrasp fails to generate meaningful grasps across these objects, which is expected since it is tailored to MANO-scale human hands and its contact representation and kinematic assumptions do not transfer well to high-DOF robotic hands.

As shown in Tab.~\ref{tab:comparison}, the conventional imitation-learning baselines (ACT-3D AND Diffusion Policy 3D) tend to reproduce trajectories close to demonstrations: as object morphology varies, they do not proactively adapt their strategy, leading to significantly degraded performance both on seen objects with poses absent from the training data and on unseen objects.
GrainGrasp, while leveraging contact cues extracted from human-hand demonstrations, is hindered by the lack of cross-demonstration consistency in the resulting contact patterns, which makes contact-conditioned pose synthesis brittle under distribution shift.
Moreover, the method is evaluated only in simulation and its contact-to-grasp stage does not explicitly enforce real-robot grasp feasibility (e.g., force-closure and frictional stability under hardware and contact uncertainties); consequently, it attains near-zero success in our real-robot protocol (Tab.~\ref{tab:comparison}), underscoring the practical difficulty of sim-to-real transfer for contact-guided grasp synthesis.
Meanwhile, CMKA's performance is limited by the layered segmentation output of SAM; as discussed in Section~\ref{subsec:Object-centric Representation Learning}, for objects that lack salient part delineation such as tomatoes and bananas, it often fails to predict correct grasp locations.
Moreover, because it executes a preprogrammed grasp approaching motion, the fingers collide with the table surface when attempting to grasp low-profile objects such as bananas and screwdrivers, thus preventing successful task completion.
% By contrast, our method achieves higher success on common objects and remains robust on previously unseen instances within the same categories. In Fig.~\ref{dex_hand}, per-finger affordance inference places contacts on grasp-functional geometry. The geometry-aligned \emph{Approach Vector} co-optimizes wrist pose and finger trajectories relative to the tabletop object, producing table-skimming entries that reliably avoid finger--table collisions, particularly beneficial for low-profile objects.

\begin{figure}[t]
    \centering
    \includegraphics[width=0.95\linewidth]{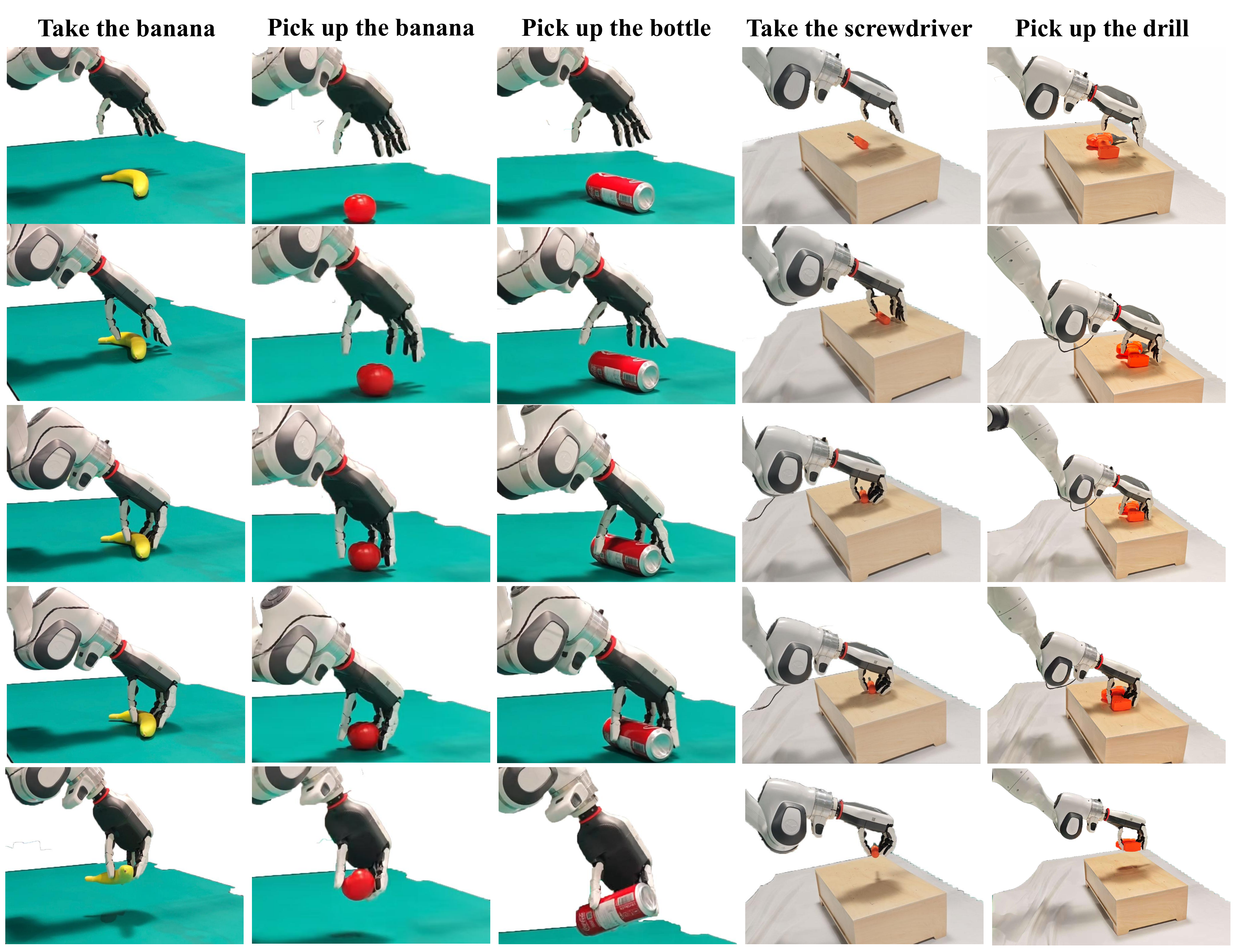}
    \caption{\textbf{Affordance-driven dexterous grasping in the real world.} Representative rollouts on everyday objects and tools; columns denote tasks (banana, bottle, tomato, screwdriver, drill).}
    \label{dex_hand}
\end{figure}

\begin{table}[t]
\centering
\caption{Performance comparison across different methods on common objects and tools manipulation tasks. Each cell reports $N{=}20$ trials per object (S = seen, U = unseen). We additionally report cross-embodiment transfer of our method on two dexterous hands.}
\label{tab:comparison}
\scriptsize
\setlength{\tabcolsep}{3pt}
\resizebox{\columnwidth}{!}{%
\begin{tabular}{lccccc}
\toprule
& \multicolumn{3}{c}{Common Objects (\%)} & \multicolumn{2}{c}{Tools (\%)} \\
\cmidrule(lr){2-4}\cmidrule(lr){5-6}
Model & Bottle (S/U) & Banana & Tomato & Screwdriver & Drill \\
\midrule
ACT-3D~\cite{gervet2023actd}                & 30/25 & 30 & 50 & 30 & 40 \\
Diffusion Policy 3D~\cite{ze20243d}         & 40/40 & 50 & 45 & 0  & 30 \\
GrainGrasp~\cite{zhao2024graingrasp}        & 0/0  & 0  & 0  & 0  & 0  \\
CMKA~\cite{yang2025multi}                   & 30/20 & 20 & 0  & 0  & 60 \\
\midrule
Ours (DexHand021)                           & \textbf{100/85} & \textbf{85} & \textbf{90}  & \textbf{60} & \textbf{70} \\
Ours (Linker Hand L20)                      & \textbf{100/80} & \textbf{90} & \textbf{100} & \textbf{70} & \textbf{90} \\
\bottomrule
\vspace{-16pt}
\end{tabular}%
}
\end{table}
\begin{figure}[t]
  \centering
  \includegraphics[width=0.95\linewidth]{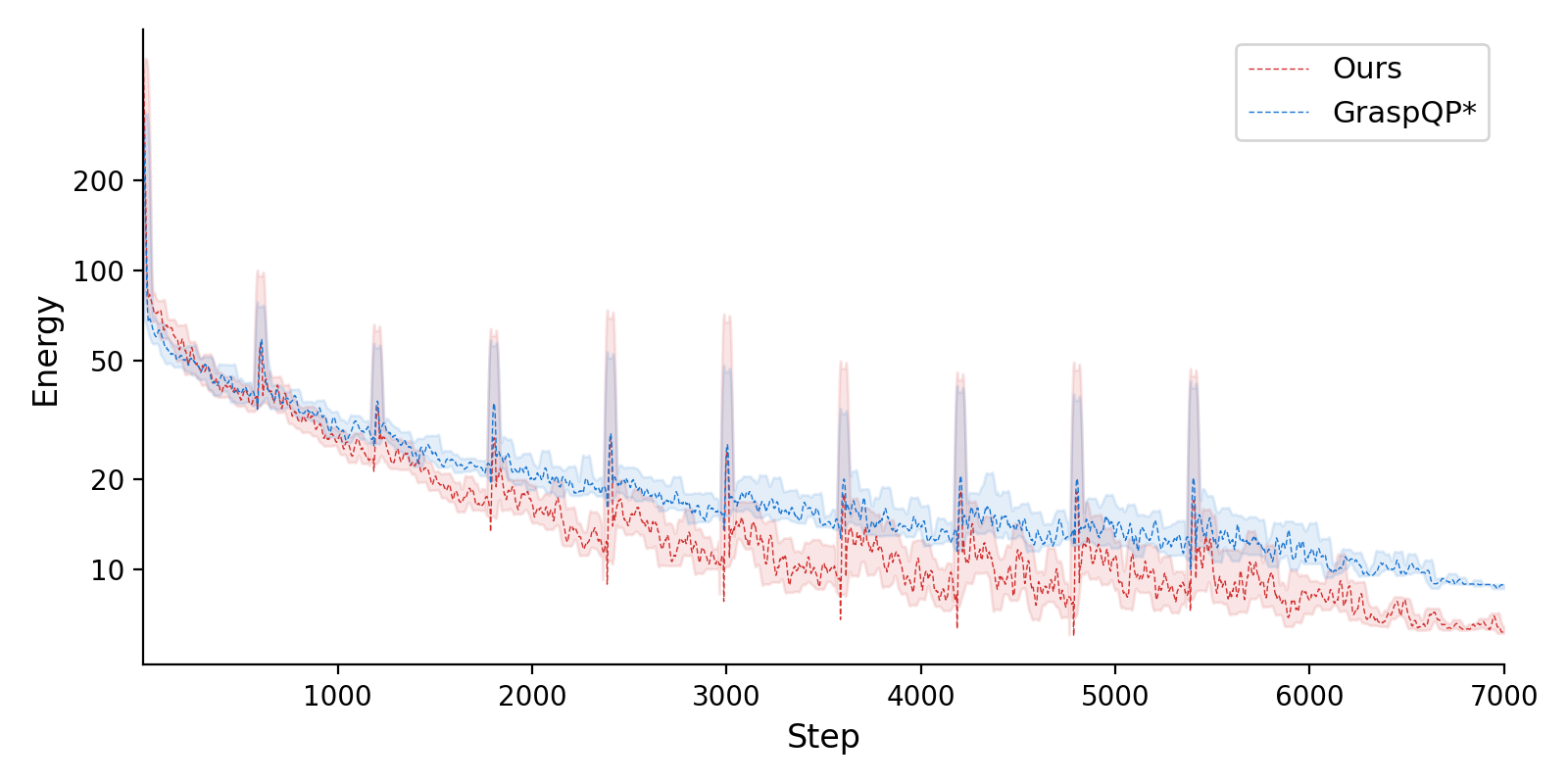}
 \caption{\textbf{Energy convergence over the full optimization.} Energy vs.\ optimization step throughout the entire refinement process. The blue curve corresponds to GraspQP* (GraspQP with the same initialization as our method), while the red curve denotes our method.}
  \label{fig:Energy_curves}
  \vspace{-10pt}
\end{figure}

% Fig.~\ref{fig:Energy_curves} plots the energy decrease after two resets. Although our objective augments the original GraspQP formulation with an additional affordance alignment term $E_{\mathrm{aff}}$, it converges faster and reaches a lower final energy than GraspQP*. We attribute this improvement to the fact that $E_{\mathrm{aff}}$ provides a dense, region-consistent geometric signal that pulls the finger contacts toward the predicted graspable surface regions, thereby reducing the search space and steering the optimizer away from poor local minima. In practice, this contact-level guidance also indirectly benefits the force-closure term $E_{\mathrm{fc}}$: by encouraging more plausible and better-situated contacts, the solver more readily finds feasible contact force distributions that balance external wrenches, leading to improved grasp quality at convergence.

Fig.~\ref{fig:Energy_curves} plots energy over the full refinement, including multiple resets. Although our objective augments the original GraspQP formulation with an additional affordance alignment term $E_{\mathrm{aff}}$, it consistently converges faster than GraspQP* after \emph{each} reset and reaches a lower final energy overall. We attribute this improvement to the fact that $E_{\mathrm{aff}}$ provides a dense, region-consistent geometric signal that pulls finger contacts toward the predicted graspable surface regions, thereby shrinking the effective search space and steering the optimizer away from poor local minima. In practice, this contact-level guidance also indirectly benefits the force-closure term $E_{\mathrm{fc}}$: by promoting more plausible and better-situated contacts, the solver more readily finds feasible contact force distributions that balance external wrenches, resulting in higher-quality grasps at convergence.

\paragraph{Cross-Embodiment Evaluation.}
Beyond the DexHand021 used in the preceding experiments, we further test the same pipeline on another dexterous hand embodiment (Linker Hand L20).We transfer the pipeline from DexHand021 (\(12{+}5\) DOFs) to Linker Hand L20 (\(16{+}5\) DOFs); perception and affordance inference remain unchanged.
As shown in Tab.~\ref{tab:comparison}, grasp success remains statistically unchanged across embodiments.
This invariance indicates that the proposed finger-specific affordance representation and geometry-aligned waypoint construction decouple grasp semantics from hand morphology, enabling deployment without hand-specific retraining.

\begin{figure}[t]
    \centering
    \includegraphics[width=0.95\linewidth]{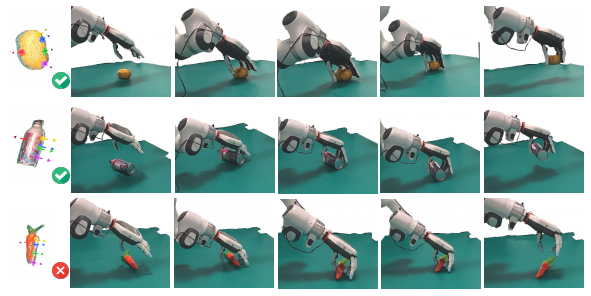}
    \caption{Zero-shot generalization on unseen objects}
    \label{fig:unseen}
    \vspace{-10pt}
\end{figure}

\section{CONCLUSION AND FUTURE WORK}
% We leverage pretrained generative models' understanding of hand--object interactions to extract both high-level object semantics and fine-grained visual cues, and fuse these with a small set of human demonstrations providing fingertip annotations. This enables learning \emph{finger-specific affordances} that unify perception and control: our representation captures not only \emph{where} to act but also \emph{how} to act. Built on this, our FSAG-based motion planning maps dexterous-hand postures onto finger-specific contacts, yielding stable functional grasps and generalization across multiple dexterous-hand embodiments. Empirically, our approach substantially outperforms contemporary methods on both affordance localization accuracy and functional grasping success rate.

We propose \emph{Finger-Specific Affordance Grounding} (FSAG), which learns per-finger contact affordances from a small set of human demonstrations. 
Our key idea is to repurpose a frozen text-to-image diffusion model as a semantic backbone and train a lightweight decoder to predict five finger-conditioned likelihood maps, bridging high-level interaction priors with fine-grained contact localization.
We lift the predicted per-finger affordance maps to object-surface regions and \emph{couple} them with physically grounded grasp refinement, jointly optimizing contact feasibility and finger–region consistency to produce stable, human-intuitive grasps.
Empirically, our approach substantially outperforms contemporary methods on both affordance localization accuracy and human-intuitive grasping success rate.

\bibliographystyle{IEEEtran}
\bibliography{IEEEtranBST/IEEEexample}

\end{document}